\documentclass[11pt]{article}

\usepackage[top=1.0in, bottom=1.0in, left=1.0in, right=1.0in]{geometry}
\usepackage{amsfonts, amssymb, amsthm, amsmath}
\usepackage{booktabs, threeparttable, multirow}
\usepackage{graphicx}
\usepackage{xcolor}
\usepackage{titlesec} 
\usepackage{caption}[2007/12/23]
\usepackage{afterpage}
\usepackage{algorithm}
\usepackage{algpseudocode}
\usepackage{setspace} 
\usepackage{authblk} 

\usepackage{todonotes}
\usepackage{appendix}
\usepackage{enumerate}
\usepackage{tabularx}
\usepackage{comment}
\usepackage{adjustbox}

\makeatletter
\newcommand{\multiline}[1]{%
  \begin{tabularx}{\dimexpr\linewidth-\ALG@thistlm}[t]{@{}X@{}}
    #1
  \end{tabularx}
}
\makeatother

\newcommand{\algorithmicelsif}{\textbf{else if}}
\algdef{C}[IF]{IF}{NoThenElseIf}[1]{\algorithmicelsif\ #1}
\usepackage{hyperref}
\hypersetup{
    colorlinks=true,        
    linkcolor=blue,         
    citecolor=blue,         
    urlcolor=blue,          
    pdfborder={0 0 0}       
}

\usepackage[backend=biber,style=apa,uniquename=false]{biblatex} 
\DeclareLanguageMapping{english}{english-apa} 
\AtEveryBibitem{\clearfield{url}} 
\AtEveryBibitem{\clearfield{doi}} 

\newtheorem{definition}{Definition}

\newtheorem{theorem}{Theorem}

\makeatletter
\def\BState{\State\hskip-\ALG@thistlm}
\makeatother

\title{\textbf{\LARGE \textcolor{darkblue}{A Novel Hybrid Heuristic–Reinforcement Learning Optimization Approach for a Class of\\ Railcar Shunting Problems}}}

\author[a]{Ruonan Zhao}
\author[a]{Joseph Geunes\thanks{Corresponding author: \texttt{geunes@tamu.edu}.}}

\affil[a]{Wm Michael Barnes '64 Department of Industrial and Systems Engineering, Texas A\&M University}

\definecolor{darkblue}{RGB}{0, 51, 102} 

\titleformat{\section}
  {\color{darkblue}\normalfont\Large\bfseries} 
  {\thesection}{1em}{} 

\titleformat{\subsection}
  {\color{darkblue}\normalfont\large\bfseries} 
  {\thesubsection}{1em}{} 

\titleformat{\subsubsection}
  {\color{darkblue}\normalfont\normalsize\bfseries} 
  {\thesubsubsection}{1em}{} 

\begin{document}
\date{}
\maketitle

\begin{abstract}
Railcar shunting is a core planning task in freight railyards, where yard planners need to disassemble and reassemble groups of railcars to form outbound trains. Classification tracks with access from one side only can be considered as stack structures, where railcars are added and removed from only one end, leading to a last-in-first-out (LIFO) retrieval order. In contrast, two-sided tracks function like queue structures, allowing railcars to be added from one end and removed from the opposite end, following a first-in-first-out (FIFO) order. We consider a problem requiring assembly of multiple outbound trains using two locomotives in a railyard with two-sided classification track access. To address this combinatorially challenging problem class, we decompose the problem into two subproblems, each with one-sided classification track access and a locomotive on each side. We present a novel Hybrid Heuristic–Reinforcement Learning (HHRL) framework that integrates railway-specific heuristic solution approaches with a reinforcement learning method, specifically $Q$-learning. The proposed framework leverages methods to decrease the state-action space and guide exploration during reinforcement learning. The results of a series of numerical experiments demonstrate the efficiency and quality of the HHRL algorithm in both one-sided access, single-locomotive problems and two-sided access, two-locomotive problems. \\

\noindent \textbf{Keywords}: Railcar Shunting; Reinforcement Learning; Heuristics
\end{abstract} \hspace{10pt}

\newpage
\section{Introduction}
Rail networks serve as a fundamental component of national and international freight transportation systems, providing high capacity, energy efficient, and cost effective movement of large volumes of goods over long distances. Within these networks, freight railyards play a central role in disassembling inbound trains and assembling outbound trains through shunting operations. Shunting, also known as marshalling or switching, refers to the movement of a single railcar or a set of continuous railcars from one track to another. These procedures are often time-consuming. As reported by \textcite{bontekoning2004breakthrough}, shunting operations in Europe can account for 10–50\% of a train’s total transit time. Meanwhile, both the United States and Europe are anticipating substantial growth in rail freight volumes \parencite{USfreight2007overview,islam2016make}. These trends emphasize the need for more effective and scalable shunting optimization models to support increasingly complex operations in modern freight yards. 

This paper considers two flat-yard configurations in which locomotives execute shunting operations, namely one-sided and two-sided yards. In one-sided yards, all tracks connect to a single switch end, so railcars are handled from the same side, which induces a last-in-first-out (LIFO) access structure. Such layouts are sometimes referred to as stub configurations.  In two-sided, or through configuration yards, tracks are accessible from both ends, so railcars may be handled from either end, including being placed and retrieved from the same side or placed from one end and retrieved from the other. When railcars enter from one end and are retrieved from the opposite end, the track can support first-in-first-out (FIFO) retrieval order. This two-end accessibility increases operational flexibility, but it also introduces additional planning complexity due to the larger set of feasible moves and the need to coordinate operations at both sides. In both configurations, tracks are partitioned into \emph{classification tracks}, which temporarily store and rearrange inbound railcars, and \emph{departure tracks}, which contain assembled outbound trains ready for departure.

The shunting process typically takes place on multiple parallel tracks connected by ladder tracks at one or both ends. Railcars with different destinations are initially distributed across multiple tracks, and locomotives then execute a sequence of moves that transfer individual railcars or contiguous blocks of railcars between tracks to form the desired outbound trains. This class of operations has been studied extensively in the literature. Prior work includes mathematical programming approaches, such as mixed-integer programming (MIP) models \parencite{trainmarshalling,switchingproject,singleengine2015}, complexity analyses \parencite{KLstacksorting,EVEN1971,PSPACECompleteStacking}, and heuristic algorithms \parencite{van2022local,jaehn2016shunting,kamenga2021solution}. More recently, reinforcement learning has been explored as a model-free approach for sequential shunting decisions \parencite{salsingikar2020reinforcement,jiaru2023shunting}. 
However, much of the existing work does not consider locomotive-based shunting costs and does not address different yard layouts, such as one-sided and two-sided configurations, or possible transformations between them.

Reinforcement Learning (RL), and in particular $Q$-learning, has demonstrated effectiveness in addressing complex and sequential decision-making problems by learning policies through interaction with the environment \parencite{watkins1992q,kaelbling1996reinforcement,sutton1998reinforcement}. However, with increasing number of states and actions, the scalability of such methods becomes a major challenge due to the exponential growth of the state--action space. In railcar shunting, naive exploration can result in spending significant effort on infeasible or low-quality movements, leading to slow convergence and poor solution performance. In contrast, railway-domain heuristics can generate feasible shunting plans efficiently by utilizing operational experience and exploiting structural features of yard operations, such as priority rules and grouping strategies \parencite{shuntingreview}, where a set of railcars is treated as a single unit and moved together. These heuristic methods are generally scalable and easy to implement, but they can be myopic, and their solution quality may decline when the yard layout or the distribution of railcars changes.

To address these challenges, this paper explores two classes of shunting problems that we refer to as the One-Sided Railcar Shunting Problem (OS-RSP) and Two-Sided Railcar Shunting Problem (TS-RSP). The OS-RSP is based on the locomotive-driven shunting formulation of \textcite{switchingproject}, which we extend to two-sided yards to obtain the TS-RSP. We then present two decomposition methods for decomposing a TS–RSP instance into two coupled OS-RSP subproblems. We propose an RL model for the OS-RSP with an objective of minimizing the shunting cost incurred in relocating specific railcars to their designated outbound tracks. This RL model treats sets of consecutive railcars sharing the same destination as unified car groups, ensuring that these groups are moved together without being split. By constructing a special case
of the TS-RSP (TS-RSP-sc), and relating it to the OS-RSP, we establish the $\mathcal{NP}$-completeness of the recognition version of the TS-RSP. We then develop a Hybrid Heuristic–Reinforcement Learning (HHRL) optimization framework to solve problems ranging from small to large scales, which contains three key processes: preprocessing, Fixed $f$-group Batching, and $Q$-learning. The proposed integrated framework solves both the OS-RSP and TS-RSP efficiently while maintaining acceptable optimality gaps, as demonstrated through comparisons with a mixed-integer programming (MIP) model from the literature.

The HHRL framework proposed in this paper can also be adapted to other combinatorial optimization problems with stack structures and precedence requirements. One example is container relocation problems that occur in day-to-day operations in container terminal storage yard, where containers are stacked vertically and need to be retrieved in a prescribed order \parencite{jin2015solving,zhu2019mathematical}. 
In this setting, the crane operator needs to determine a sequence of container moves according to LIFO order. Another application arises in the steel industry, where steel slabs are stored in stacks in a slab yard and must be retrieved by bridge cranes according to a scheduled sequence \parencite{tang2010modelling}. 

The main contributions of this paper are as follows:
\begin{enumerate}
    \item We introduce two mapping functions that decompose any TS-RSP instance into two coupled OS-RSP subproblems that can be solved in parallel, while explicitly accounting for the coordination of two locomotives.

    \item We cast the OS-RSP/TS-RSP in terms amenable to RL, more specifically $Q$-learning.  The proposed approach enables flexible movement of either a single railcar or any number of consecutive railcars with an objective of optimizing locomotive-based shunting costs. This also allows flexible movement between any pair of tracks, including classification-to-classification, classification-to-departure, and departure-to-departure transfers.

    
    \item Due to the large numbers of railcars and tracks in practical freight yards, we develop an HHRL framework that integrates railway-domain heuristics with $Q$-learning to improve scalability and solution quality. The framework includes two heuristic procedures, preprocessing and fixed $f$-group batching, which can standardize arbitrary initial yard layouts, reduce the effective state–action space, and guide exploration during $Q$-learning.

    \item We provide extensive computational results on 120 instances, ranging from small to large, that demonstrate the effectiveness of the proposed HHRL approach across both one-sided and two-sided yard types and different problem scales.

    \item We study the relationship between the TS-RSP and OS-RSP by comparing their makespans, defined as the total number of time periods required to complete a shunting plan. We show that TS-RSP achieves significantly smaller makespans across instances.
\end{enumerate}

The remainder of this paper is organized as follows. Section~\ref{sec:literature} reviews related work from prior literature. Section~\ref{sec:formulation} provides formal definitions of the OS-RSP and TS-RSP.  Section \ref{sec:SolutionMethod} then discusses a heuristic solution approach that first decomposes an instance of the TS-RSP into two OS-RSP subproblems (Section \ref{sec:decom}) and then applies the $Q$-learning framework presented in Section \ref{sec:$Q$-learningmodel} to solve each of the resulting OS--RSP subproblems. Section~\ref{sec:hhrl} describes the proposed HHRL framework, and Section~\ref{sec:results} reports computational experiments that evaluate its performance. Section~\ref{sec:conclusion} concludes the paper and outlines directions for future research.

\section{Literature Review}
\label{sec:literature}
In flat railyards, shunting operations are executed by locomotives, whereas in hump yards, railcars are shunted primarily by gravity. Researchers have developed a wide range of solution approaches for shunting operations, which are also referred to as sorting, marshaling, or switching in the literature, and substantial progress has been made over the past few decades. Terminology varies across studies and application settings. For example, railcars may be called cars or wagons, and one-sided tracks are often modeled as stacks and referred to as stub yards due to their LIFO access property \parencite{wolfhagen2017train}. \textcite{shuntingreview} provide an excellent and comprehensive overview of the operations decisions in shunting yards, as well as a thorough review of prior research. As in many other combinatorial optimization problems, existing solution strategies generally fall into exact methods and heuristic algorithms.

The literature considers shunting problems at three decision levels: strategic, tactical, and operational.
Strategic studies address long-term design and capacity questions, such as the number, location, and layout of yards, as well as the availability of tracks, switches, and engines.
\textcite{belovsevic2018variable} propose a binary integer programming formulation to simultaneously optimize a classification schedule and railyard design by determining the structural layout and track capacity in the yard. Due to the computational complexity of the problem, they develop a variable neighborhood search heuristic. \textcite{zhang2017simu} propose a simulation model for analyzing railyard capacity that enables estimating dwell times as a function of several design and operational factors (e.g., number of engines, yard volume, assemble rate, and humping speed).  

At the tactical level, researchers primarily focus on timetable planning and network routing decisions, along with the design of priority rules and blocking policies that specify how railcars are grouped into destination-based blocks to reduce dwell time in the yard.
\textcite{carey2003scheduling} propose a scheduling heuristic that assigns each train to a station platform while avoiding conflicts with other trains. \textcite{caprara2011solution} study the train platforming problem, which assigns arriving and departing trains to station platforms while accounting for their arrival and departure times and operational constraints (e.g., at most one train can occupy a platform at any time). They formulate the problem as an integer linear programming model. A key difference is that \textcite{caprara2011solution} assume a periodic setting in which trains follow fixed arrival and departure times that repeat each day, whereas \textcite{carey2003scheduling} do not impose this daily fixed-timetable assumption. \textcite{dewilde2013robust} propose a heuristic approach to update an existing train timetable by jointly considering routing decisions and platform assignments.
 \textcite{zhu2014scheduled} further develop an integrated framework based on a three-layer network that captures key tactical planning activities, including train service selection and scheduling, railcar blocking decisions, and operations performed on cars in yards. They formulate the resulting problem as an MIP model with an objective of minimizing total system cost. 

Operational studies, which are the focus of this paper, address short-term, detailed decisions regarding the movement, rearrangement, and sorting of railcars within a yard to assemble outbound trains. These models must consider practical constraints that are fixed in the short term such as track capacities and locomotive routing limitations. A key challenge at this level is to find an efficient sequence of shunting operations that minimizes total cost, total processing time, and/or the number of classification tracks needed to assemble outbound trains. Researchers have approached these problems using various methods including MIP, dynamic programming (DP) and domain-specific heuristics. Several studies investigate the \emph{Train Marshaling Problem} posed by \textcite{trainmarshalling}. In the particular problem class considered by \textcite{trainmarshalling}, a single train enters a yard on a track at one end of the yard that splits into multiple parallel tracks.  These parallel tracks subsequently come together at the other end of the yard and feed a departure track. Switching the railcars onto different parallel tracks on one end and pulling from different parallel tracks on the other end permits sorting cars into a different order from the order in which they entered (in this case, railcars exit parallel sorting tracks in FIFO order).  

\textcite{PSPACECompleteStacking} use an approximate DP method to minimize the number of sorting moves, while not allowing movements of multiple cars at the same time. \textcite{falsafain2019novel} seek to minimize the number of classification tracks used for shunting by developing a DP approach, whose worst-case running time grows exponentially with the number of destinations but scales proportionally with the number of railcars. \textcite{bosi2024optimal} formulate a mixed-integer nonlinear programming model with multiple objectives, including minimizing the set of cars selected for shunting, the operational cost of shunting, and the delays induced by shunting operations. \textcite{Lubbecke2005} study railcar management in in-plant rail networks. They divide adjacent tracks into regions and decompose the problem into two subproblems: allocating railcars of each type from each region to meet the demand request, and executing the within-region retrieval of the assigned railcars. The second subproblem involves shunting operations, and they propose a MIP model to minimize shunting cost. However, their formulation is restricted to a single outbound track and adopts a depth-based cost structure in which the shunting (retrieval) cost is proportional to a railcar’s position along that track. In addition, their model limit car transfers only in one direction, which is from a classification track to the outbound track.
In our problem, we allow flexible shunting movements between any pair of tracks, including classification-to-classification, departure-to-departure, and transfers between classification and departure tracks in either direction.


While most prior studies rely on traditional OR techniques or heuristic approaches, recent research has started to explore machine learning, particularly RL, to solve shunting yard problems across strategic, tactical, and operational decision levels. For instance, \textcite{vsemrov2016reinforcement} introduce RL for railway rescheduling when an initial schdule is disrupted, for example due to accidents or equipment failures. A $Q$-learning method is used to reduce train delays. \textcite{ScalableRL_IEEE} introduce an algorithm for generating a timetable in railway lines based on an RL approach. 

At the operational level, however, relatively few studies have applied RL specifically to locomo\-tive-driven shunting. As noted by \textcite{shuntingreview}, the detailed movement of freight railcars from classification tracks to departure tracks by locomotives within a flat yard remains an open and underexplored problem. To the best of our knowledge, \textcite{hirashima2011reinforcement} presents the closest related study to ours, proposing a $Q$-learning approach that minimizes the total distance traveled by a locomotive while forming a single outbound train. 
Their model, however, differs from ours in several key respects. First, their model only focuses on creating one outbound train, whereas our model offers the flexibility of assembling multiple outbound trains. Second, their approach requires relocating all railcars and lining them onto a main track to assemble the outbound train, while our model offers the flexibility of only moving specific railcars to form the outbound trains.
Additionally, they consider a one-sided yard layout, where railcars can only be accessed from a single end of each track. Our work addresses both one-sided and two-sided yards and introduces mapping functions that transform a two-sided instance into two coupled one-sided subproblems. Building on this representation, we propose a novel Hybrid Heuristic--Reinforcement Learning (HHRL) framework. The power of the HHRL framework is that it offers considerable versatility: (i) it allows moving any number of railcars simultaneously between any pair of tracks, including classification-to-classification, departure-to-departure, and transfers between classification and departure tracks in either direction; (ii) it supports forming any number of outbound trains rather than a single outbound train, with the number of trains limited only by the available departure tracks; 
and (iii) it accommodates different types of railcars, including railcars with destinations that are needed to form outbound trains, and railcars without destination (e.g., empty cars) that only need to remain on classification tracks.

\section{Problem Definition and TS-RSP Transformation}
\label{sec:formulation}
This section provides a detailed description of the One-Sided Railcar Shunting Problem (OS-RSP) and the Two-Sided Railcar Shunting Problem (TS-RSP), and then discusses two techniques for decomposing a TS-RSP instance into two coupled OS-RSP subproblems. Before introducing the formal definitions, we first define common terminology and notation used in both problems. In a railyard, \emph{classification tracks} temporarily store and rearrange railcars during the shunting process, while \emph{departure tracks} serve as assembly locations for outbound trains with specific destinations. We allow shunting movements between any pair of tracks, including classification-to-classification, departure-to-departure, and transfers between classification and departure tracks in either direction. A railcar \emph{group} is defined as a contiguous set of railcars that share the same destination, and individual cars within a group are always moved together as a single unit. 

Let $S$ denote the set of tracks, partitioned into a subset of classification tracks $S_C$ and a subset of departure tracks $S_D$, with $S_C \cup S_D =S$ and $S_C\cap S_D=\emptyset$. Each track $s\in S$ has an associated track length $L_s$.  Let $R$ denote the set of railcar groups, comprised of the subsets $R_D$ and $R_C$, where $R_D$ are groups with designated destinations in $S_D$ and $R_C$ are groups without designated destinations that may be placed on any track in $S_C$ (with $R_C \cup R_D =R$ and $R_C\cap R_D=\emptyset$). Each group $r\in R$ has length $l_r$, with $l_r \ge 1$, and an associated destination track $d(r)\in S_D$ for $r\in R_D$, and $d(r)\in S_C$ for $r\in R_C$. Each group has an initial location specified by a mapping $f:R\to S_C \times \mathbb{Z}_{+}$, which assigns group $r$ to its initial classification track and position on that track. Positions on a track are indexed by depth from the switch end: position 1 is the car closest to the switch end, position 2 is immediately behind it, and so on. 

\subsection{OS-RSP definition}
In a one-sided flat yard, all tracks are accessible to a locomotive at a single end, denoted as the \emph{Switch End}, while the opposite side is known as the \emph{Dead End}. Shunting moves correspond to the use of a locomotive to transfer a railcar group (or multiple groups) from one track to another via the connecting, or ladder tracks. Because each track is accessible from only one end, car groups are accessed in a LIFO order on each track.

A shunting move transfers a group or multiple groups from track $i$ to track $j$ via the switch end. For any ordered pair $(i,j)\in S\times S$ with $i\neq j$, let $c_{ij}\ge 0$ denote the shunting cost associated with moving a group or multiple groups from $i$ to $j$ via the switch end. This cost is designed to capture the locomotive effort, including labor and diesel emissions, and can be modeled in particular as a function of the travel distance between tracks, although our model permits any cost that is independent of the number of cars moved between the tracks.

\begin{figure}[htbp!]
\begin{center}
\includegraphics[scale=0.18]{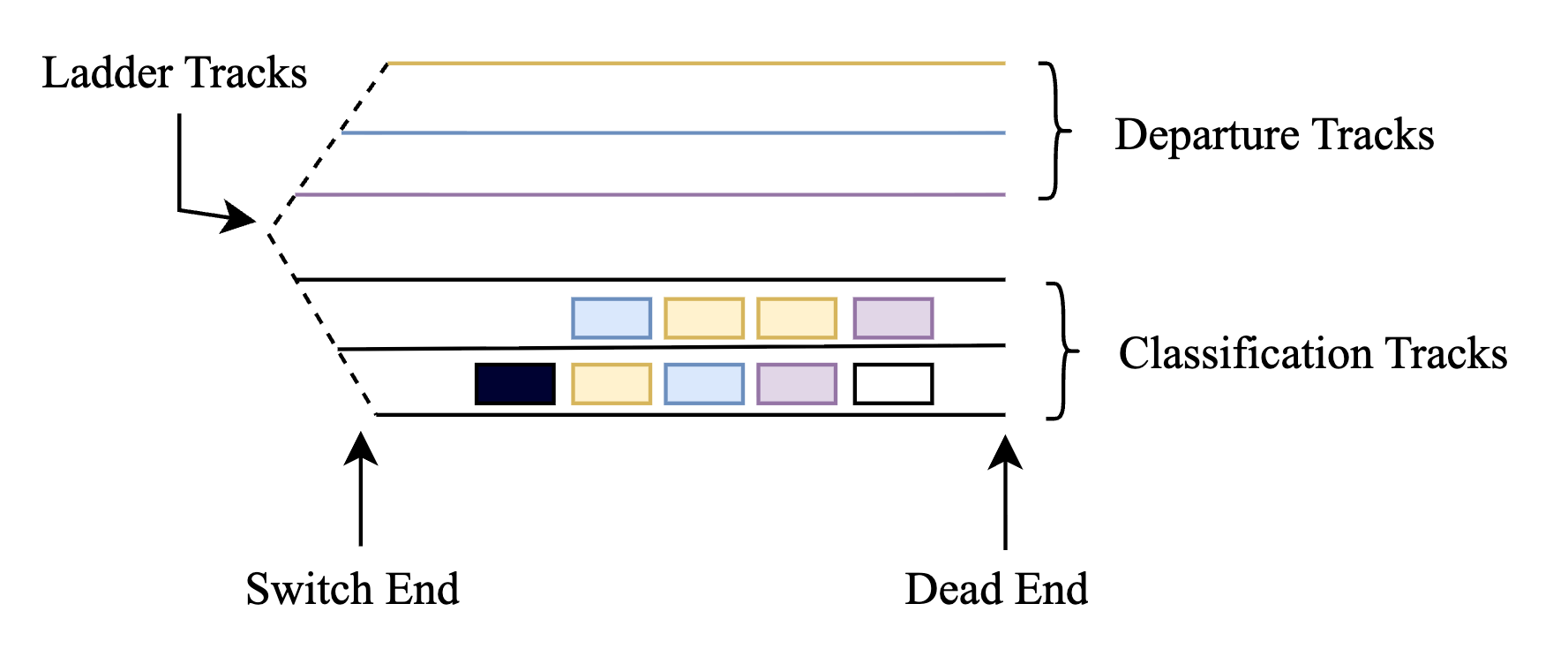}
\caption{One-sided railyard layout example.} \label{OS-RSP}
\end{center}
\end{figure}

Figure \ref{OS-RSP} illustrates a sample one-sided railyard layout with three departure tracks, three classification tracks, two ladder tracks, eight railcars, and one locomotive. In the figure, railcars are depicted as rectangles, with the black railcar representing the locomotive. Classification tracks are denoted by black horizontal lines, while departure tracks are color-coded (yellow, blue, and purple) to indicate different destinations. Additionally, each railcar is also color-coded to correspond with its designated departure track. White railcars, however, do not have destinations, and can be assigned or moved to any of the classification tracks. In this one-sided railyard, tracks are interconnected on the left side by ladder tracks, allowing railcars to be accessed exclusively from this switch end and subsequently moved to any other track as needed. Instead of managing individual railcars, we manage railcar groups. As shown in the figure, the higher classification track contains three groups: a blue group, a yellow group, and a purple group. The blue and purple groups each consist of a single car, while the yellow group consists of the two adjacent yellow cars. During shunting, the feasible shunting movements for this track from the switch end are to move (i) the blue car alone, (ii) the blue car together with the entire yellow group, or (iii) the blue car together with the yellow group and the purple group. Similarly, the four different color-coded railcars on the lowest classification track are considered
as four separate groups, each consisting of only one railcar.

Building upon the above operational framework, the OS-RSP can be defined as follows.

\begin{definition}(OS--RSP)
Given the one-sided track set $S= S_D\cup S_C$, railcar group set $R= R_D\cup R_C$, group length $l_r$ for each $r\in R$, initial location mapping $f:R\to S_C \times \mathbb{Z}_{+}$, and shunting cost $c_{ij}$ for any pair of tracks $(i,j)\in S\times S$ with $i\neq j$, the OS-RSP seeks a sequence of shunting moves that delivers all groups to their respective destination tracks, in any order, at a minimum the total shunting cost.
\end{definition}

\textcite{switchingproject} show that the OS--RSP is an $\mathcal{NP}$-hard optimization problem.  We next generalize this problem to one in which classification tracks are accessible by a locomotive from either side, i.e., no dead-end exists.

\subsection{TS-RSP definition}
We next present the TS-RSP, which generalizes OS-RSP by allowing shunting from both ends of the yard using two locomotives operating simultaneously. The goal of the TS-RSP is to determine an optimal sequence of railcar
movements within a two-sided railyard that minimizes the total shunting cost required to assemble
outbound trains. Unlike the OS-RSP, where all relocations occur from a single switch end, TS-RSP allows
shunting from both ends of the yard using two locomotives operating simultaneously.

\begin{figure}[h]
\centering
\includegraphics[width=0.75\textwidth]{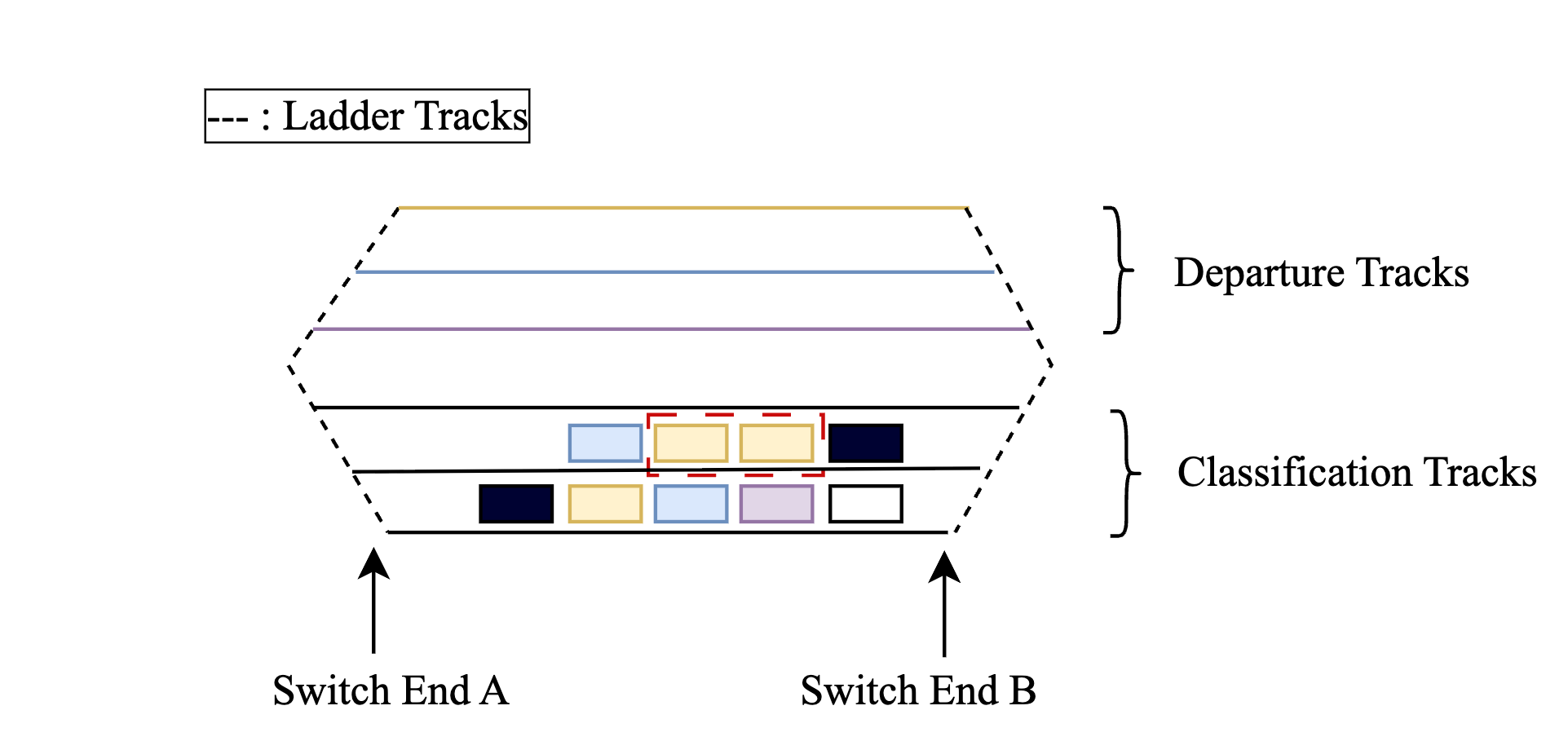}
\caption{Two-sided railyard layout example}
\label{TS-RSP}
\end{figure}

The yard layout includes ladder tracks, classification tracks, and departure tracks. Figure \ref{TS-RSP} shows an example of symmetric railyard layout. We use the same visualization scheme as in the OS-RSP figures (Figure \ref{OS-RSP}) to represent departure tracks, classification tracks, railcars, and locomotives. The main difference is that, in the TS-RSP, each track connects to ladder tracks at both ends, allowing locomotives to access classification and departure tracks from either end. The two switch ends, denoted as \emph{Switch End A} and \emph{Switch End B}, allow simultaneous shunting operations using two locomotives and facilitate the transfer of railcar groups between tracks at either end.  This structure permits accessing cars in FIFO or LIFO order (or using a combination of the two).

Railcars are initially located on classification tracks, with some requiring movement to specific departure tracks and others without predetermined destinations. Shunting decisions are made at the group level and we never split a group, similarly to OS-RSP. For example, the two yellow railcars highlighted in Figure~\ref{TS-RSP} are always moved as a single unit. The shunting cost is modeled as a function of the distance the locomotive travels between tracks. For any ordered pair of tracks $(i,j)$, we denote by $c_{ij}^A$ (respectively \ $c_{ij}^B$) the cost of moving a group or multiple groups from track $i$ to track $j$ via Switch End A (respectively, Switch End B). In a symmetric yard layout as shown in Figure~\ref{TS-RSP}, the costs $c_{ij}^A$ and $c_{ij}^B$ are typically equal. However, in yard configurations with asymmetric geometry, these costs may differ even for the same pair of tracks $(i,j)$.

Based on the operational framework described above, the TS-RSP is defined as follows.

\begin{definition}(TS--RSP)
Given the two-sided track set $S=S_D \cup S_C$, railcar group set $R = R_D \cup R_C$, group length $l_r$ for each $r\in R$, initial location mapping $f:R\to S_C \times \mathbb{Z}_{+}$, and shunting costs $c_{ij}^A$ and $c_{ij}^B$ for shunting any number of groups from track $i$ to track $j$ via Switch End $A$ and $B$, respectively, the TS-RSP seeks a sequence of shunting operations that delivers all groups to their respective destination tracks, in any order, while minimizing the total shunting cost.
\end{definition}

Because the special case of the TS-RSP in which $c_{ij}^B=\infty$ for all $(i,j)\in S\times S$ reduces to the OS--RSP, the complexity results for the OS-RSP from \textcite{switchingproject} imply that the TS-RSP is an $\mathcal{NP}$-hard optimization problem as well. Because of this, we propose solving the TS-RSP by decomposing it into two separate OS-RSP instances, and solving each instance heuristically using an RL approach. The following section develops our proposed solution method.

\section{RL-Based Heuristic Solution Method}\label{sec:SolutionMethod}
As mentioned earlier, we first propose two techniques for decomposing an instance of the TS-RSP into two separate instances of the OS-RSP. For any instance of the OS-RSP, Section \ref{sec:$Q$-learningmodel} presents a $Q$-learning approach for solving the problem.  Because the state space for practical problem sizes becomes enormous, Section \ref{sec:hhrl} describes a heuristic approach for its implementation that enables solving practical problem sizes in acceptable computing time.

\subsection{Decomposing the TS-RSP into coupled OS-RSP subproblems}\label{sec:decom}
As shown in \textcite{switchingproject}, the OS-RSP is a large-scale combinatorial optimization problem, and solving the model exactly requires long computing times, even for small problem instances.  While the heuristic method they proposed enables solving medium-size problems, the required solution time for the proposed heuristic grows rapidly, prohibiting solution of larger problem instances within a reasonable amount of computing time.  Because the TS-RSP is at least as hard as the OS-RSP, we propose a hybrid heuristic solution technique that incorporates problem decomposition and RL. This overall solution approach is developed throughout the remainder of this section.  

We first decompose each TS-RSP instance into two coupled OS-RSP subproblems, one associated with switch end $A$ and one associated with switch end $B$. The key idea is to partition, on each track, the ordered sequence of groups into two subsets, with one locomotive performing shunting operations only for groups assigned to switch end $A$ and the other shunting only the groups assigned to switch end $B$. This partition is implemented by inducing an internal dead end between the two subsets of groups on each track. Under this construction, each locomotive shunts only its assigned side and does not cross the internal dead end, which avoids conflicts between the two ends and allows the two one-sided subproblems to be solved in parallel.

When the number of groups on a classification track is an even number, we can assign half of the groups to each end, splitting them in the middle.  For example, if there are $n$ groups on a classification track and $n$ is an even number, then we assign the $\frac{n}{2}$ groups closest to switch end $A$ to subproblem $A$ and the remaining groups to subproblem $B$.  When $n$ is odd, we assign $\left\lceil \frac{n}{2} \right\rceil$ groups to one end, and $\left\lfloor \frac{n}{2} \right\rfloor$ groups to the other end.  To do this, we consider two mappings. The first, \emph{A-Preferential Split (APS)}, assigns the extra group to switch end \(A\). Alternatively, one can define an equivalent variant that always assigns the extra group to switch end \(B\). The second, \emph{Rotating Odd-Balance Split (ROBS)}, alternates the assignment of the extra group between switch ends \(A\) and \(B\) across successive classification tracks. APS provides a baseline that is easy to implement and analyze. This mapping is appropriate when one switch end is treated as the default shunting side due to operational preference. In contrast, ROBS is designed to balance the assignment across the two ends by alternating which end receives the extra group on odd tracks, which can balance shunting operations on both  sides.

In both mappings, the two-sided yard is partitioned into two coupled one-sided yards that can operate in parallel. Formally, for each track $i$ with $h_i$ groups, we define the allocations $(A_i,B_i)$ as follows. Here, $A_i$ denotes the numbers of groups assigned to switch end $A$, i.e., the consecutive $A_i$ groups closest to switch end $A$ on track $i$; $B_i$ is the number of remaining groups assigned to switch end $B$, with $A_i+B_i=h_i$.  We next summarize these two mapping approaches.

\begin{itemize}
    \item \textbf{APS.} For a track $i$ with $h_i$ ordered groups, APS assigns the extra group to switch end $A$ when $h_i$ is odd and splits evenly otherwise:
    \[
    A_i=\left\lceil \frac{h_i}{2}\right\rceil,\qquad
    B_i=\left\lfloor \frac{h_i}{2}\right\rfloor,
    \]
    equivalently,
    \[
    (A_i,B_i)=
    \begin{cases}
    \left(\dfrac{h_i}{2},\,\dfrac{h_i}{2}\right), & \text{if } h_i \text{ is even},\\[6pt]
    \left(\dfrac{h_i+1}{2},\,\dfrac{h_i-1}{2}\right), & \text{if } h_i \text{ is odd}.
    \end{cases}
    \]

    \item \textbf{ROBS.}
    ROBS splits evenly when $h_i$ is even. When $h_i$ is odd, it alternates which switch end receives the extra group across successive odd tracks.
    Let $\tau\in\{0,1\}$ be an odd-track indicator initialized to $0$. Define
    \[
    (A_i,B_i)=
    \begin{cases}
    \left(\dfrac{h_i}{2},\, \dfrac{h_i}{2}\right), & \text{if } h_i \text{ is even},\\[8pt]
    \left(\left\lceil \dfrac{h_i}{2}\right\rceil,\; \left\lfloor \dfrac{h_i}{2}\right\rfloor\right), 
    & \text{if } h_i \text{ is odd and } \tau=0,\\[8pt]
    \left(\left\lfloor \dfrac{h_i}{2}\right\rfloor,\; \left\lceil \dfrac{h_i}{2}\right\rceil\right), 
    & \text{if } h_i \text{ is odd and } \tau=1.
    \end{cases}
    \]
    After processing an odd $h_i$, update $\tau \leftarrow 1-\tau$; otherwise leave $\tau$ unchanged.
\end{itemize}

Either of these mappings decomposes the TS-RSP into two coupled OS-RSP subproblems that can be solved in parallel, referred to as subproblem $A$ and subproblem $B$. Note that because switch end $B$ lies on the opposite side of the yard, subproblem $B$ is a mirrored version of the standard OS-RSP as illustrated in Figure \ref{OS-RSP}; equivalently, it can be represented as an OS-RSP after reversing the yard orientation so that the switch end is on the left. For example, consider a single track $i$ whose groups, listed from switch end $A$ (left) to switch end $B$ (right), are \([g_1,\; g_2,\; g_3,\; g_4,\; g_5]\). Suppose a split yields $(A_i,B_i)=(3,2)$. Then subproblem $A$ contains the three groups closest to switch end $A$, namely \([g_1,\; g_2,\; g_3]\), and the subproblem $B$ contains the two groups closest to switch end $B$, namely \([g_4,\; g_5]\). Subproblem $A$ is already in the standard OS-RSP layout with the switch end on the left, while subproblem $B$ is reversed, with the switch end on the right. To express subproblem $B$ in the standard OS-RSP orientation with the switch end on the left, we reverse the group ordering on the track, yielding \([g_5,\; g_4]\), which is exactly the sequence moving away from the switch end. Hence, solving subproblem $B$ is equivalent to solving a standard OS-RSP with the switch end on the left and the dead end on the right. 

\subsection{$Q$-learning Model Formulations}
\label{sec:$Q$-learningmodel}
As previously discussed, both OS-RSP and TS-RSP aim to find a shunting sequence that relocates all car groups to their respective destination tracks while minimizing total operational cost. Moreover, each TS-RSP instance can be decomposed into two coupled OS-RSP subproblems, subproblems $A$ and $B$, which  can be solved in parallel using one locomotive for each subproblem. To optimize the shunting sequence, we develop an RL approach based on $Q$-learning for the OS-RSP and, using the proposed decomposition, apply this to both subproblems to solve the TS-RSP.

Before describing the RL formulation, we introduce the following assumptions and definitions. To keep track of the total time associated with a proposed solution, we assume that moving any subset of railcar groups from one track to another (i.e., one shunting move) requires one time period. Recall that the \emph{makespan} is the total number of time periods required to complete a shunting plan. A smaller makespan indicates faster completion of all required moves and thus more time-efficient yard operations. For the OS-RSP, because a single locomotive is used, the makespan equals the total number of shunting steps. For the TS-RSP, two locomotives operate in parallel from the two switch ends; therefore, the makespan is the maximum between the number of shunting steps for subproblems $A$ and $B$, i.e., the time until both sides finish. For each group $r$ and time period $t$, let $P_{rt}$ denote the position of group $r$ on its current track, which equals one plus the number of groups between $r$ and that track's switch end at time $t$. Thus, $P_{rt}=1$ for the group closest to the switch end, and the group position index increases by one as we move away from the switch end. The \emph{head group} of a track is the group at the switch end in position $1$ (i.e., the group with the smallest index on that track), while the \emph{tail group} of a track is the group with the largest index on that track.

\begin{figure}[h]
\centering
\includegraphics[width=0.56\textwidth]{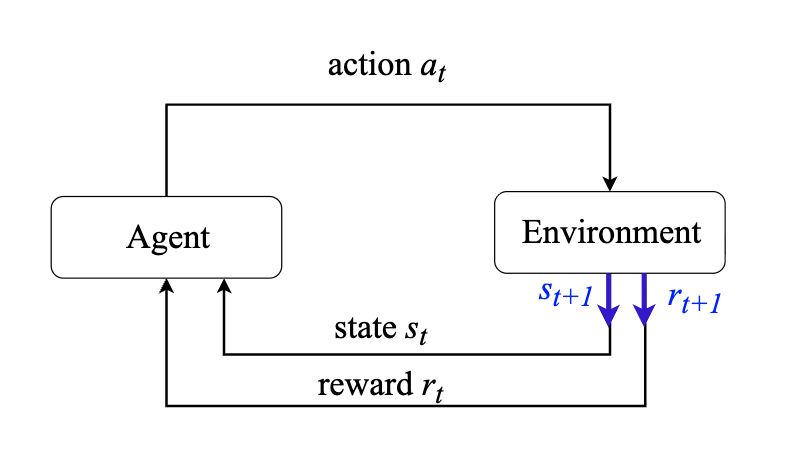}
\caption{The general setting for RL}
\label{RLSET}
\end{figure}

RL models sequential decision-making in which a policy selects actions based on observed states, the environment transitions in response, and the objective is to find a policy with maximum expected cumulative reward. The general setting for RL is shown in Figure \ref{RLSET}. At each decision step \(t\), the agent receives the current state \(s_t\) and responses by choosing an action \(a_t\) according to the environment's decision policy. The agent executes the selected action, inducing an environment transition from state \(s_t\) to the next state \(s_{t+1}\) and generating a reward \(r_{t+1}\). This reward reflects each action's effectiveness: actions that improve the environment yield higher rewards, while less effective actions result in lower or even negative rewards.
To maximize cumulative rewards, the agent balances exploration, which selects new actions to improve its knowledge of their outcomes in the environment, and exploitation, which uses current value estimates of the environment to choose the best action \parencite{kaelbling1996reinforcement}. This general setting allows policy learning for the agent from repeated trials and errors, without requiring prior knowledge of the environment’s transition dynamics. 

In the railcar shunting setting, the agent corresponds to the yardmaster or planning system that decides which railcar groups to move and to which track. The environment is the railyard system: the configuration of classification, departure and ladder tracks, the ordered railcar groups on each track, and the associated movement cost matrix. Other fundamental components are defined as follows.

\textit{1) States.}
A state at time \(t\), denoted \(s_t \in \mathcal{S}\), where \(\mathcal{S}\) denotes the set of possible states, is encoded as the tuple
\[
  s_t=\bigl(k_1,k_2,\dots,k_{|S|}\bigr),
\]
where \(k_i\) is an ordered list of the groups from switch end to dead end on track \(i\) , \(\forall  i \in S\). The initial state \(s_0\) represents the original configuration of the railway system. A terminal state is any configuration that simultaneously satisfies the following two conditions:  
(i) every group \(r \in R_D\) has been moved to its designated departure track \(d(r) \in S_D\); 
and (ii) every group \(r \in R_C\) is placed on some classification track in \(S_C\).  

\textit{2) Actions.}
Given \(s_t\), an action is taken in time \(t\), denoted as \(a_t \in \mathcal{A}(s_t)\), where
\(\mathcal{A}(s_t)\) represents the set of all possible actions from state $s_t$. Each element of \(\mathcal{A}(s_t)\) represents the movement of one or more contiguous railcar groups from the switch end of a source track to the switch end of a receiving track during the corresponding time period. The source track can be any track in \(S\) that currently holds at least one railcar group, while the receiving track can be any other track in \(S\), including either a classification track or a departure track. Formally, we represent the action at time \(t\)  by \(a_t=(i,m,j)\) with \(i,j \in S \times S, \text{and }i\neq j\). Here, \(m\) denotes the number of contiguous railcar groups (starting at the head group) to be shunted from track \(i\), where \(1 \le m \le h_i\) and \(h_i\) is the total number of groups currently on track \(i\). The selected \(m\) groups are moved together as one block, maintaining their original sequence on the receiving track.

\textit{3) Reward function.}
After taking action $a_t$ from state $s_t$, a transition occurs to a new state $s_{t+1}$ and a reward $r_{t+1}$ is realized that provides feedback on the chosen action (we supress the dependence of $r_{t+1}$ on the state and action for notational convenience). The reward function is constructed to match the OS-RSP objective of minimizing total shunting cost while delivering all groups to their designated destination tracks. We combine an immediate learning signal with a delayed completion signal. Specifically, when we execute a shunting move that transfers one or more groups from track $i$ to track $j$, we receive an immediate reward equal to the negative shunting cost, $-c_{ij}$. In addition, to encourage completion of the shunting plan, a completion bonus $B>0$ is awarded when the resulting state is a terminal state, i.e., all groups are on their designated destination tracks. Thus, we define the corresponding reward function as:
\[
  r_{t+1} \;=\; -\,c_{ij} \;+\;
  \begin{cases}
    B, & \text{if } s_{t+1} \text{ is terminal},\\
    0, & \text{otherwise},
  \end{cases}
\]
assuming the action taken corresponds to a shunting move from track $i$ to $j$.  

$Q$-learning \parencite{sutton1998reinforcement} defines the agent’s expected reward from a decision policy through the state-action value function \(Q(s_t, a_t)\), i.e., this function estimates the benefit when action $a_t$ is executed if in state $s_t$ and an optimal policy is subsequently followed. The agent typically begins with no prior knowledge about the quality associated with actions, and we therefore initialize all \(Q\)-values to zero. Through the learning process, the value function \(Q\) is updated based on the realized reward observed in the training phase, as described by the following update rule \parencite{sutton1998reinforcement}:
\begin{equation}\label{eq: Qlearning}
  Q(s_t,a_t) \leftarrow Q(s_t,a_t)
    + \alpha \Bigl[r_{t+1} + \gamma \max_{a_{t+1}}Q(s_{t+1},a_{t+1}) - Q(s_t,a_t)\Bigr].
\end{equation}
In Equation \eqref{eq: Qlearning}, the parameter \(\alpha \in (0, 1]\) corresponds to a so-called learning rate, \(\gamma \in [0, 1)\) corresponds to a discount rate, and \(r_{t+1}\) is the immediate reward received when  action \(a_t\) is chosen in state \(s_t\). The learning rate \(\alpha\) determines the speed with which $Q$-values change based on new experience. A low learning rate implies that the agent updates its $Q$-values more slowly, while a high learning rate indicates the agent quickly adapts to new information, placing more weight on more recent experiences. The discount rate \(\gamma \in [0, 1)\) controls the relative values of future rewards to immediate rewards. As \(\gamma\) approaches 0, the agent focuses on maximizing immediate rewards, which means the agent is short-sighted; as it nears 1, future rewards are given greater importance in the agent's evaluation of $Q$-values.

The $Q$-learning Model involves both a training (learning) phase and an inference (optimization) phase, where training iteratively improves the $Q$-values through trial and error, and the final optimized $Q$-table values then determine the action that maximizes reward for the current state. During training, the agent employs an \(\varepsilon\)-greedy strategy to balance exploration and exploitation, with minimum exploration rate $\varepsilon_{\min}$. With probability $\varepsilon$ the agent randomly selects an action in \(\mathcal{A}(s_t)\) in order to explore states that may not otherwise be selected.  Otherwise (with probability $1-\varepsilon$) the agent chooses the action $a_t$ that maximizes $Q(s_t,a_t)$ when in state $s_t$, exploiting the collective knowledge obtained up to that point.  After each episode (i.e., a complete training run that starts from the initial yard configuration and continues until reaching a terminal state), the exploration rate \(\varepsilon\) may be reduced based on a decay schedule according to
\begin{equation}\label{eq: explorerate}
  \varepsilon \; =\; \max \left\{\varepsilon_{\min},\, \varepsilon \cdot \varepsilon_{\mathrm{decay}}\right\},
\end{equation}
where the parameter \(\varepsilon_{\mathrm{decay}} \in (0, 1)\) controls the rate of decay. This mechanism ensures that the agent is sufficiently likely to explore during early episodes, but gradually shifts to exploitation as learning progresses. The detailed training algorithm is provided in Algorithm \ref{alg:qlearning}. 

In the algorithm, values in the $Q$-table are initialized to zero for each $(s, a)$ pair upon its first visit, i.e., when a state $s$ is first visited, we initialize \(Q(s, a)=0\) for all feasible actions $a \in A(s)$.  We then run $M$ training episodes to learn $Q(s_t,a_t)$ values, where $M$ is a training parameter. In each episode, the agent starts from the selected initial yard configuration $s_0$ and repeatedly selects a feasible shunting action using the \(\varepsilon\)-greedy rule. Equation \eqref{eq: Qlearning} is applied to update $Q$-values, and an episode ends when we reach a terminal state.  Upon reaching a terminal state, the cumulative reward of the episode $G_e$ is recorded accordingly. After each episode, the exploration rate $\varepsilon$ is updated according to \eqref{eq: explorerate}. 

\begin{algorithm}[htbp]
\caption{$Q$-Learning Training Algorithm for OS-RSP}\label{alg:qlearning}
\begin{algorithmic}[1]
\State Initialize $Q(s, a) \gets 0$ for any $(s,a)$ when $s$ is first visited and $a\in\mathcal{A}(s)$
\State Initialize $G_e \gets 0$ for each episode
\For{$e = 1$ to $M$} \Comment{Loop over episodes}
  \State $t \gets 0$.
  \State Set $s_0$ to the chosen initial state.
  \While{$s_t$ is nonterminal}
    \State Select $a_t$ from $\mathcal{A}(s_t)$ using $\varepsilon$-greedy policy.
    \State Apply transition and reach $s_{t+1}$.
    \State Observe reward $r_{t+1}$.
    \State Update $Q$-value by \(
      Q(s_t, a_t) \gets Q(s_t, a_t) + \alpha \Bigl[r_{t+1} + \gamma \max_{a_{t+1}} Q(s_{t+1}, a_{t+1}) - Q(s_t, a_t)\Bigr]\).
    \State Accumulate reward: $G_e \gets G_e + r_{t+1}$.
    \State Update state: $s_t \gets s_{t+1}$.
    \State $t \gets t + 1$.
  \EndWhile
  \State Update exploration rate:
  \(
    \varepsilon \gets \max\left(\varepsilon_{\min},\, \varepsilon \cdot \varepsilon_{\mathrm{decay}}\right).
  \)
\EndFor
\end{algorithmic}
\end{algorithm}

After training, we obtain a $Q$-table that we then apply as an optimization policy for each test instance. Starting from the initial state, at each time period, we select a feasible action using a greedy policy, i.e., by choosing the action $a_t$ in period $t$ according to
\begin{equation}\label{eq: greedy_action}
a_t \;\in\; \arg\max_{a\in\mathcal{A}(s_t)} Q(s_t,a).
\end{equation}
We then apply the corresponding action, and repeat until a terminal state is reached. If multiple actions have the same $Q$-value in any given state, then the agent selects one of these randomly. We define the \emph{$Q$-learning cost} as the accumulated shunting cost until reaching a terminal state, and we denote by $T_Q$ the makespan of the corresponding sequence of actions.
\subsection{HHRL Approach}\label{sec:hhrl}
Section \ref{sec:formulation} defined the OS-RSP and TS-RSP and presented methods to decompose a TS-RSP instance into two coupled OS-RSP subproblems. Section \ref{sec:$Q$-learningmodel} introduced the $Q$-learning model for the OS-RSP, which facilitates decision-making in shunting yards while optimizing total shunting cost. However, the total number of feasible state-action pairs grows rapidly and significantly with the number of railcar groups and tracks, leading to computational challenges in the training phase. Furthermore, given the problem’s complexity and the scale of practical problem instances, implementing the $Q$-learning model can be computationally prohibitive, and finding an optimal solution may be impossible within acceptable computing time.  Therefore, we next propose an HHRL algorithm designed to generate high-quality solutions efficiently for real-world problem sizes while maintaining an acceptable optimality gap.

\begin{figure}[h]
\centering
\includegraphics[width=0.95\textwidth]{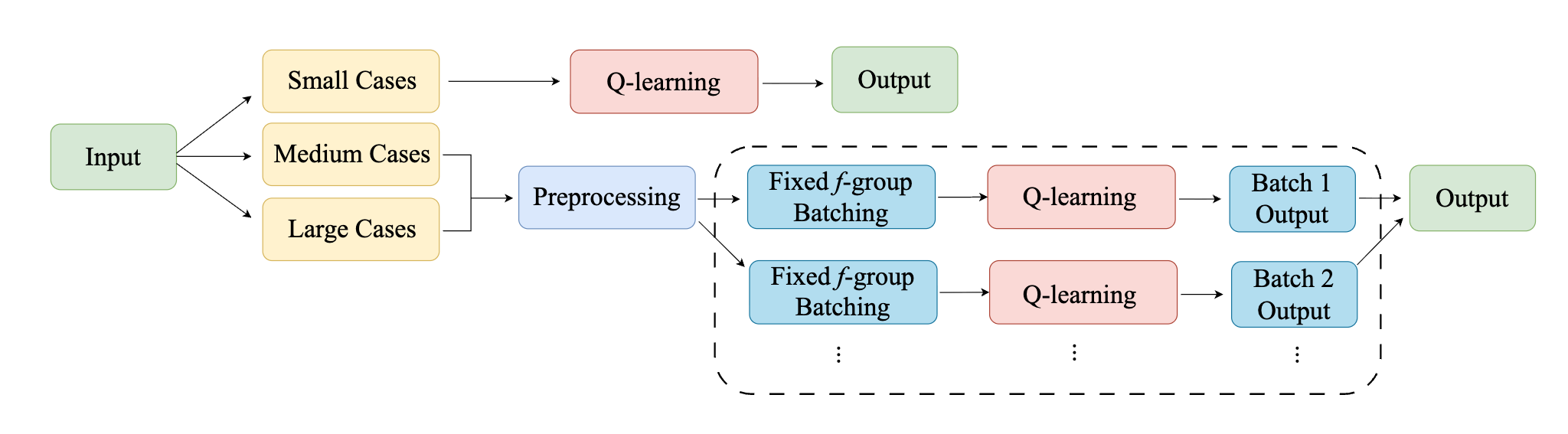}
\caption{HHRL framework}
\label{HHRLfr}
\end{figure}

 The HHRL framework is shown in Figure \ref{HHRLfr}, which consists of three main processes: preprocessing, fixed $f$-group Batching, and $Q$-learning. Preprocessing, discussed in Section \ref{sec:preprocess}, involves a set of simple steps applied to an initial yard layout that permit reducing the number of possible state-action pairs.  The fixed $f$-group batching process, described in Section \ref{sec:fgroup}, decomposes each problem into a set of smaller subproblems, each of which has a manageable state space.  
 
\begin{figure}[h]
\centering
\includegraphics[width=0.6\textwidth]{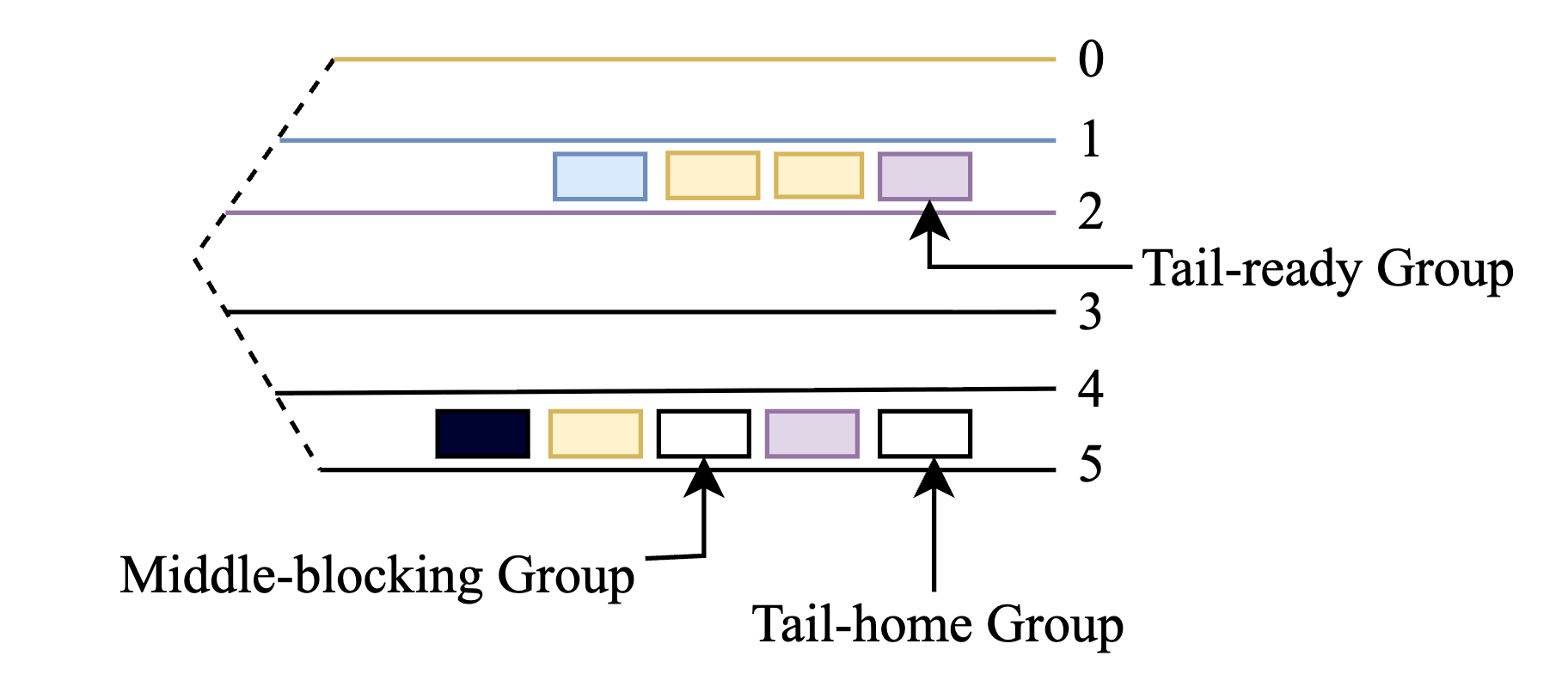}
\caption{Group type illustration}
\label{gtype}
\end{figure}

Before describing these processes, we introduce the following definitions. Note that these definitions are based on an instance of the OS-RSP, i.e., the shunting problem with one switch end and one dead end.  In this context, we define a \emph{merge} as the process of combining two railcar groups that share the same destination into a single group.  Recall that the position indices of railcar groups on a track start at 1 for the head group and increase by one as we move toward the dead end. We defined the tail group as the group with the largest index on that track. Consider a track $j$ with ordered railcar groups $(r_1,\ldots,r_x)$ from switch end to dead end, where group $r_i$ has destination \( d(r_i) \in S_C \cup S_D \), and $r_x$ denotes the tail group on track $j$. If $d(r_x)=j$ with $j\in S_D$, we call $r_x$ a \emph{tail-ready group}; if $d(r_x)=j$ with $j\in S_C$, we call $r_x$ a \emph{tail-home group}. A group $r_i$ is called a \emph{middle-blocking group} if \( d(r_i) \in S_C \), and it is neither the head group nor the tail group on that track. Figure \ref{gtype} presents an example of these group types. 

Let \(S=\{0,1,\dots,n-1\}\) denote the track set, which we partition into \(S_D=\{0,\dots,k_0-1\}\) and \(S_C=\{k_0,\dots,n-1\}\). We refer to $k_0$ as the \emph{top classification track}, which is also the lowest indexed classification track, and thus the classification track closest to the departure tracks. For example, in Figure~\ref{gtype}, track 3 is the top classification track. With a slight abuse of notation, we denote \(d_h(i)\) as the designated destination track associated with the head group on track \(i \in S\).  Because our preprocessing routine involves merging head groups in relatively close proximity to one another, we define \(\delta\in\mathbb{Z}_{\ge 0} \) as a threshold value on the difference between track indices.  We then define the track pair $(i,j)$ with $i<j$ as a \emph{head-pair} if \( d_h(i)=d_h(j) \) and \( j-i<\delta\). More specifically, we distinguish two subclasses of head-pairs:
\begin{itemize}
  \item \textbf{U-pair} (unique destination): \(d_h(i)=d_h(j)\in S_D\), i.e., both head groups on track $i$ and $j$ are assigned to the same departure track.
  \item \textbf{F-pair} (flexible destination): \(d_h(i)=d_h(j)\in S_C\), i.e., both head groups on track $i$ and $j$ may be sent to any classification track.
\end{itemize}

\subsection{Preprocessing}\label{sec:preprocess}
Preprocessing standardizes different initial yard configurations through a structured sequence of shunting moves presented in Algorithm \ref{alg:Preprocessing}. The accumulated shunting cost resulting from these moves is defined as the \emph{preprocessing cost}. The goal is to transform an arbitrary initial state into a standardized layout, while decreasing the number of possible state-action pairs. The algorithm consists of five phases. 

\begin{algorithm}[t]
\small
\caption{Preprocessing Algorithm}\label{alg:Preprocessing}
\begin{algorithmic}[1]
\State Initialize preprocessing cost \(C_p \gets 0\).

\State \textbf{Phase 1. Delete tail-ready group and tail-home
group}
    \For{each nonempty $j \in S$}
        \If{there is tail-ready group or tail-home group}
            \State Delete them
        \EndIf
    \EndFor
\Statex
\State \textbf{Phase 2: Merge of head groups in the head-pair }
\While{true}
  \State Let $\mathcal{P}$ be the list of tuples $(i,j,\kappa)$, where $i,j$ are the head-pair in state $s$, and $\kappa \in \{U, F\} $ is the subclass of the head-pairs.
  \If{$\mathcal{P} = \emptyset$} \State \textbf{break} \EndIf
  \State \textbf{sort} $\mathcal{P}$ by key $(j-i,\, i)$ in nondecreasing order
  \State $(i,j,\kappa) \gets \mathcal{P}[1]$ \Comment{first element after sorting}
  \If{$\kappa = U$}
     \State Move the head group on track $j$ to $i$, and obtain state $s'$.
  \Else \Comment{$\kappa = F$}
     \State Move the head group on track $i$ to $j$, and obtain state $s'$.
  \EndIf
  \State \(C_p \gets j-i\)
  \State $s \gets s' $
\EndWhile
\Statex

\State \textbf{Phase 3. Stepwise moves to the top classification track $k_0$}
\State Let $L=[y_1, y_2, \dots, y_m]$ be the increasing list of nonempty classification tracks
\For{$v = m, m-1, \dots, 2$}
    \State $j \gets y_v$
    \State $i \gets y_{v-1}$
    \State Move the cars on track $j$ to track $i$
    \State \(C_p \gets C_p + j-i\)
\EndFor

\If{$y_1 \neq k_0$}
    \State Move the cars on track $y_1$ to the top classification track $k_0$
    \State \(C_p \gets C_p + y_1 - k_0\)
\EndIf
\Statex
\State \textbf{Phase 4. Clear groups $r \in R_C$ from the top classification track $k_0$}
    \If{the head group $r$ on the top classification track $k_0$ belong to $R_C$}
        \State Move $r$ from $k_0$ to track $k_0+1$
        \State \(C_p \gets C_p + 1\)
    \EndIf
    \While{a middle-blocking group exists}
        \State Move all cars up to and including the first middle-blocking group to track $k_0+1$
        \State \(C_p \gets C_p + 1\)
        \State Move colored cars from track $k_0+1$ to track $k_0$
        \State \(C_p \gets C_p + 1\)
    \EndWhile
\Statex

\State \textbf{Phase 5: Delete specific classification tracks}
    \State Delete tracks with an index greater than $k_0$.
\end{algorithmic}
\end{algorithm}

\paragraph{Phase 1} The first phase deletes all tail-ready and tail-home groups. Since these groups are already on their designated destination tracks and do not block any other groups, they do not require further shunting. Removing them from the problem reduces the number of groups in the system and thus reduces the state and action space for subsequent steps.

\paragraph{Phase 2} The second phase merges head groups on identified head-pairs to further reduce the total number of groups. Because we may have multiple head-pairs with the same destination, we sort these pairs and prioritize merges with the smallest index distance, since such merges typically incur lower locomotive shunting cost. When multiple head-pairs have the same distance value $(j-i)$, we start with the pair with the smallest track index $i$. For each selected head-pair $(i,j)$, the merge direction depends on its type. For a U-pair, the two head groups share a fixed destination in $S_D$, so we move the head group on track $j$ to track $i$, which positions the merged group closer to its designated departure track. In contrast, for an F-pair, the two head groups have flexible destinations in $S_C$, so we move the head group on track $i$ to track $j$, which places the merged group farther from the departure tracks and aligns with
the intended use of classification tracks. In both cases, two head groups sharing the same destination are merged into a single larger group, which is never split in subsequent shunting operations.

\paragraph{Phase 3} The third phase consolidates all remaining railcar groups onto the top classification track \(k_0\) to form a single train consisting of groups in \(R_C\) and \(R_D\).\footnote{For problem instances in which $\sum_{r \in R_D} l_r>L_{k_0}$, we consolidate groups onto the top $n$ classification tracks such that sufficient track length exists on these $n$ tracks to accommodate all groups in $R_D$.  In other words, we define the top $n$ tracks as an extended track $k_0$.} Recall that the top classification track has the smallest index among all classification tracks. These moves are implemented sequentially over the set of nonempty classification tracks \(L=[y_1,\ldots,y_m]\): starting from the largest indexed nonempty track $v$, we move the entire set of railcar groups on track \(y_v\) to track \(y_{v-1}\) for \(v=m,m-1,\ldots,2\). This sequential aggregation moves groups that are initially scattered across multiple classification tracks onto the top classification track to form a complete train.

\paragraph{Phase 4} After forming a single train on track $k_0$, we wish to remove any groups in $R_C$ (groups without a destination) located on this track. These groups obstruct the movements needed to arrange and dispatch departing groups (i.e., those in $R_D$). We therefore move groups in $R_C$ from $k_0$ to another classification track, so that track $k_0$ contains only groups in $R_D$. Because each group in $R_C$ has a flexible destination that can be any classification track, relocating it to any other classification track places it on an admissible destination. These relocated groups can then be treated as resolved and removed from further consideration, reducing the overall number of groups and shrinking the subsequent state and action space.  Because tail-home groups are removed in Phase~1, any group in $R_C$ on track $k_0$ can only appear at the head (position $1$) or in a middle position (i.e., neither head nor tail). We clear these groups in two steps. First, if the head group on track $k_0$ belongs to $R_C$, we move this head group directly from $k_0$ to the adjacent classification track $k_0+1$. Next, we eliminate any middle-blocking groups on $k_0$ (i.e., groups in $R_C$ whose positions are neither the head nor the tail). Whenever such a group exists, we transfer all groups that are to the left of the first middle-blocking group—including the middle-blocking group itself—to track $k_0+1$. Any $R_D$ groups that were moved to $k_0+1$ as part of this transfer are then moved back from $k_0+1$ to $k_0$. Repeating this procedure until no middle-blocking group remains ensures that all remaining groups on $k_0$ belong to $R_D$.  

\paragraph{Phase 5} The last preprocessing phase deletes all classification tracks with an index higher than $k_0$. After this step, the yard is reduced to a
standardized layout in which $k_0$ is the only remaining classification track, and all unresolved groups (i.e., groups in $R_D$) are located on
this track.

\subsection{Fixed $f$-Group Batching}\label{sec:fgroup}
After preprocessing, any arbitrary initial yard configuration is transformed into a standardized layout, i.e., one in which all railcar groups in $R_D$ are located on track $k_0$, which does not contain any middle-blocking groups or head groups from $R_C$. The fixed $f$-group batching process shown in Algorithm \ref{alg:batching} then decomposes this standardized state into a sequence of batches by scanning the top classification track $k_0$ from the switch end toward the dead end and partitioning the ordered car groups into consecutive batches of size $f$ (with the final batch possibly smaller when the total number of groups on the top classification track $k_0$ is not a multiple of $f$). For example, the first batch includes the first $f$ groups nearest to the switch end, the second batch includes the next $f$ groups, and so on until all groups are assigned. Each batch contains at most $f$ car groups, while preserving all original tracks as in the standardized state. This decomposition does not incur any shunting cost, since it does not involve any shunting moves. 

Within each batch, we apply the $Q$-learning procedure and train the agent on that batch independently and consecutively. We start with batch 1: we train the agent on batch 1 and then apply the resulting policy to execute the shunting moves such that all groups in this batch reach their destination tracks. We then move to batch 2, repeat the training and execution steps, and continue until all batches are processed. During this training, we restrict the allowable shunting actions to occur only between the classification track $k_0$ and the destination tracks of
the groups contained in the current batch. This restriction substantially reduces the number of state–action pairs that must be explored and stored during $Q$-learning. Overall, this decomposition allows the agent to train on batches and obtain corresponding $Q$-tables or policies for each batch, enabling scalable training and execution without exhaustively exploring the full state–action space. 

\begin{algorithm}[H]
\caption{Fixed $f$-group Batching Algorithm} \label{alg:batching}
\begin{algorithmic}[1]
\State $\mathcal{B} \gets [\,]$ \Comment{list of batches}
\State $r \gets 1$ \Comment{batch index}
\While{the top classification track $k_0$ in state $s$ is not empty}
  \State identify the leftmost $f$ group $f^{(r)}$ on $k_0$
  \State \textbf{output:} groups contained in batching $r$ $:= f^{(r)}$
  \State append $f^{(r)}$ to $\mathcal{B}$
  \State update $s$ by removing $f^{(r)}$ from $k_0$
  \State $r \gets r+1$
\EndWhile
\State \Return $\mathcal{B}$ \Comment{$\mathcal{B} = [f^{(1)},f^{(2)},\dots]$}
\end{algorithmic}
\end{algorithm}

We next present computational results evaluating the performance of the HHRL solution approach on instances of both the OS-RSP and the TS-RSP.

\section{Computational Results}
\label{sec:results}
We conducted a set of computational experiments to benchmark how the proposed HHRL algorithm performs on OS-RSP and TS-RSP problem instances of varying size. We ran all instances on a machine with an Apple M3 Pro chip and 18 GB of memory. For OS-RSP instances, we benchmark the quality of solutions and computing time of the HHRL approach against the MIP model and heuristic approach provided in \cite{switchingproject}; the MIP is solved using Gurobi Optimizer 11.0.3 with a Python 3.9.19 implementation. For TS-RSP instances, we compare the corresponding results under the two different proposed decomposition mapping functions: APS and ROBS. Overall, these experiments provide insights into how the HHRL approach performs on both the OS-RSP and TS-RSP.

According to the Railroad Classification Yard Technology Manual published by the U.S.\ Department of Transportation \parencite{wong1981railroad}, flat yards, in which cars are pulled or pushed by locomotives, are designed with combined classification, receiving, and departure tracks, and large flat yards may have up to 40 tracks in total. To distinguish problem sizes, we classify flat yard scales jointly by the total number of tracks (including classification and departure tracks) and the total number of railcar groups, as summarized in Table~\ref{tab:yardscale}.

\begin{table}[htbp]
\centering
\caption{Problem scale by number of car groups and tracks}
\label{tab:yardscale}
\begin{tabular}{l|ccc}
\hline
\textbf{Number of Railcar groups \textbackslash\ Tracks} & \textbf{0--10} & \textbf{10--20} & \textbf{20--40} \\
\hline
\textbf{0--10} & Small  & Small  & Medium \\
\textbf{10--20} & Small  & Medium & Large  \\
\textbf{20--40} & Medium & Large  & Large  \\
\hline
\end{tabular}
\end{table}

A total of 60 test cases were randomly generated for OR-RSP, including 20 small-, 20 medium-, and 20 large-scale cases. Each OS-RSP instance was then converted into a corresponding TS-RSP instance by modifying the yard structure to include two switch ends while keeping the same set of railcar groups and their initial positions. This procedure yields 120 instances in total: 60 OS-RSP instances and 60 TS-RSP instances, with 20 small-, 20 medium-, and 20 large-scale cases in each set. Table \ref{tab:experimentparameters} presents the parameter settings used to generate test instances at different scales, where the discrete uniform distribution is denoted by $\mathcal{U}(l, u)$  with support over the integers \(l, l + 1, \dots, u\). Table \ref{tab:experimentparameters2} contains additional problem parameter rules applied across all problem instances at each scale.  Note that our computational tests assign $c_{ij}$ values based on the differences in track indices, which are proportional to inter-track distances.  Thus, for example, if inter-track distances are equally spaced with an inter-track distance of one unit, then the objective measures the distance traveled by the locomotive.
\begin{table}[htbp]
\centering
\footnotesize
\caption{Scale-dependent parameter settings for problem instance generation}
\label{tab:experimentparameters}
\begin{threeparttable}
\begin{tabularx}{\textwidth}{@{}X c c c c@{}}
\toprule
\textbf{Parameter} & \textbf{Symbol} & \textbf{Small} & \textbf{Medium} & \textbf{Large} \\
Number of total tracks & $|S|$ & $\mathcal{U}(4,10)$ & $\mathcal{U}(10,40)$ & $\mathcal{U}(10,40)$ \\
Number of departure tracks & $|S_D|$ &
$\mathcal{U}\!\bigl(2,\min(|S|-2,4)\bigr)$ &
$\mathcal{U}\!\bigl(5,\min(|S|-2,7)\bigr)$ &
$\mathcal{U}\!\bigl(8,\min(|S|-2,10)\bigr)$ \\
Number of total car groups & $|R|$ & $\mathcal{U}(2,20)$ & $\mathcal{U}(2,40)$ & $\mathcal{U}(10,40)$ \\
\bottomrule
\end{tabularx}
\end{threeparttable}
\end{table}

\begin{table}[htbp]
\centering
\footnotesize
\caption{Scale-independent parameter settings for problem instance generation}
\label{tab:experimentparameters2}
\begin{threeparttable}
\begin{tabular}{l c c}
\toprule
\textbf{Parameter} & \textbf{Symbol} & \textbf{Value(s)}\\
Number of classification tracks & $|S_C|$ & $|S|-|S_D|$ \\
Number of total car groups in $R_C$ & $|R_C|$ & $\mathcal{U}(0,\min(|R|-1,10))$ \\
Number of total car groups in $R_D$ & $|R_D|$ & $|R|-|R_D|$ \\
Length of car group $r$, $\forall r \in R$ & $l_r$ & 1 \\
Cost of moving car groups from track $i$ to $j$, $\forall i,j \in S, j \neq i$& $c_{ij}$ &  $|i-j|$ \\
Head-pair index threshold & \(\delta\) &  3 \\
\bottomrule
\end{tabular}
\end{threeparttable}
\end{table}

\begin{figure}[htbp!]
\begin{center}
\includegraphics[scale=0.62]{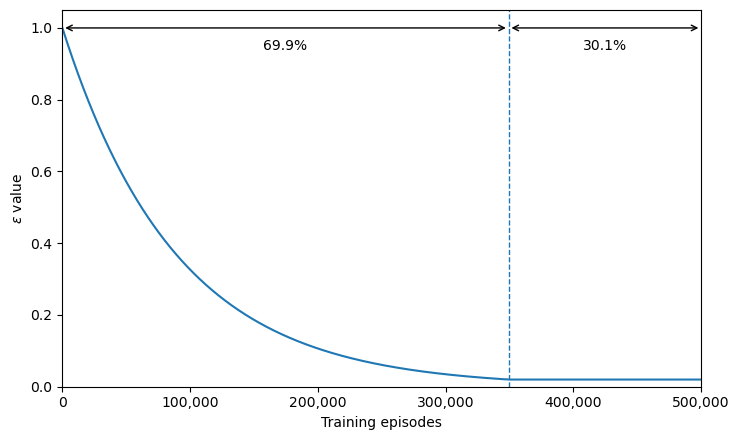}
\caption{Exploration rate decay over training episodes} \label{exploration_rate}
\end{center}
\end{figure}

We next describe the $Q$-learning parameter settings used in our experiments. The number of training episodes used was $M=500,000$ for all cases. The learning rate $\alpha$ and discount rate $\gamma$ were set at $0.1$ and rate $0.99$, respectively. The choice of action was determined by using an $\varepsilon$-greedy policy with initial exploration rate $\varepsilon=1.0$.  This exploration rate decays gradually based on the application of Equation \eqref{eq: explorerate} at each episode.  The values used for $\varepsilon_{\mathrm{decay}}$ and $\varepsilon_{\min}$ were
$0.9999888$ and $0.02$, respectively. 
With these settings, about 70\% of the training episodes have an exploration rate greater than 0.02, as shown in Figure~\ref{exploration_rate}, which illustrates a high exploration rate during early episodes.

The completion bonus component of the reward function varies by instance size. Specifically, we set $B=15$ for small cases, $B=30$ for medium cases, and $B=60$ for large cases. For small cases, we solve each instance directly using the $Q$-learning Algorithm \ref{alg:Preprocessing}, and the corresponding total shunting cost (referred to as the HHRL cost) equals the $Q$-learning cost. For medium and large cases, we first apply preprocessing and then use the fixed $f$-group batching procedure to decompose the resulting standardized state into smaller substates in each batch; the successive batch subproblems are then solved sequentially using $Q$-learning. Thus, the HHRL cost equals the preprocessing cost plus the sum of the $Q$-learning costs over all batches. The makespan for each case is also reported to capture operational efficiency gains from the TS-RSP over the OS-RSP.

To evaluate the proposed HHRL algorithm for OS-RSP instances, we compare its performance against the MIP model and the adaptive railcar grouping dynamic programming (ARG-DP) heuristic reported in \textcite{switchingproject}. For the OS-RSP, we compare the solution quality and running time of the HHRL approach to those for the MIP model and the ARG-DP heuristic. We measure solution quality using the optimality gap, computed as the HHRL cost minus the optimal cost, divided by the optimal cost, for all problems which Gurobi was able to solve to optimality. 

Table \ref{tab:mip_argdp_hhrl} summarizes average performance measures for each solution method for the different problem scales. The column labeled “Obj.” denotes the objective function value, which corresponds to the total shunting cost, “Time (s)” denotes running time in seconds, and “Gap (\%)” denotes the percentage optimality gap. The objective is measured in cost units, which, in our computational tests, serve as a proxy for the total distance traveled by the locomotive (assuming equally spaced inter-track distances). Note that for the 20 medium-scale cases, the MIP solver was only able to find solutions for 6 instances within the 12-hour time limit. Therefore, we report the MIP results separately for these 6 solvable cases. For these cases, the ARG-DP heuristic attains an average optimality gap of 2.30\% with an average running time of 688.10 s, while the HHRL method achieves a 0\% gap in 13.01 seconds. For the remaining 14 medium-scale cases, neither the MIP nor ARG-DP heuristic produces solutions within 12 hours, while the HHRL method successfully generates solutions for all instances with reasonable running times. Overall, the results show that the HHRL method consistently delivers quality solutions across all problem sizes. It remains effective even as the problem size increases, producing feasible results where the MIP model and existing heuristic methods are unable to return any solutions within the time limit. This makes the HHRL method a practical and robust approach for solving complex OS-RSP instances.

\begin{table}[htbp]
    \centering
    \footnotesize
    \caption{Performance comparison of MIP, ARG-DP, and HHRL for OS-RSP}
    \label{tab:mip_argdp_hhrl}
    \resizebox{\textwidth}{!}{%
    \begin{tabular}{lccccccccc}
        \toprule
        & \multicolumn{2}{c}{MIP} & \multicolumn{3}{c}{ARG-DP} & \multicolumn{3}{c}{HHRL} \\
        \cmidrule(lr){2-3} \cmidrule(lr){4-6} \cmidrule(lr){7-9}
        Scale & Obj. & Time (s) & Obj. & Opt. Gap (\%) & Time (s) & Obj. & Opt. Gap (\%) & Time (s) \\
        \midrule
        Small (20 cases) & 6.80 & 21.12  & 7.15 & 4.78 & 1.65 & 7.05 & 3.05 & 117.15  \\
        Medium (6 cases) & 19   & 14.14  & 19.5 & 2.30  & 688.10 & 19 & 0 & 13.01  \\
        Medium (14 cases) & --   & --     & --   & --   & -- & 33.50 & -- & 178.68  \\
        Large (20 cases)  & --   & --     & --   & --   & -- & 61.20 & -- & 332.68 \\
        \bottomrule
        \multicolumn{9}{l}{\footnotesize \textit{Note:} “--” indicates no results were obtained within the 12-hour time limit.}
    \end{tabular}%
    }
\end{table}

We next present average performance measures under the two decomposition mapping functions, APS and ROBS, for handling the TS-RSP in Table \ref{tab:aps_robs_makespan}. In addition to the objective value (Obj.) and running time in seconds (Time), we also report the makespan. This metric captures operational time efficiency: a smaller makespan corresponds to faster completion of the required shunting operations. Note that the optimization phase in $Q$-learning takes under 0.001 seconds; therefore, the running time of the HHRL model does not include this time. Overall, both APS and ROBS scale well with two-sided problem sizes in terms of running time; for all cases, the largest average running time is 302.9 s. The largest average objective function value is 73.7 units, which corresponds to the number of inter-track distance units traveled by both locomotives. These results indicate that the HHRL approach can solve larger-scale instances while maintaining acceptable computation times.
The ROBS decomposition approach consistently yields smaller average makespan values across all problem sizes, suggesting more efficient shunting move sequences. However, the ROBS approach also shows slightly higher average objective values (total shunting cost) compared to APS, indicating a trade-off between minimizing makespan and minimizing overall shunting cost.

\begin{table}[htbp]
\centering
\scriptsize
\setlength{\tabcolsep}{9pt} 
\caption{Performance comparison of APS and ROBS under HHRL for TS--RSP}
\label{tab:aps_robs_makespan}
\begin{tabular}{lcccccc}
\toprule
& \multicolumn{3}{c}{HHRL-APS} & \multicolumn{3}{c}{HHRL-ROBS} \\
\cmidrule(lr){2-4} \cmidrule(lr){5-7}
Scale & Obj. & Makespan & Time (s) & Obj. & Makespan & Time (s) \\
\midrule
Small (20)  & 8.0  & 3.75  & 37.70  & 8.1  & 3.50  & 36.72 \\
Medium (20) & 32.1 & 10.60 & 107.72 & 36.1 & 8.75  & 118.20 \\
Large (20)  & 72.1 & 21.35 & 293.16 & 73.7 & 17.15 & 302.90 \\
\bottomrule
\end{tabular}
\end{table}

To further test our hypothesis that the mean makespan for the TS-RSP is significantly smaller than that for the OS-RSP, we compare their makespan values in Table~\ref{tab:ms_comparison} and report the corresponding $p$-values for a statistical test on the mean differences. The \emph{makespan reduction} is calculated as 
\(
\left(1-\frac{\text{makespan of TS--RSP}}{\text{makespan of OS--RSP}}\right)\times 100\%
\), which represents the percentage decrease in the total number of time periods required to complete a shunting plan. In the table, we denote this metric by MS Red.\ (\%). Across all scales, the TS-RSP yields smaller average makespans than the OS-RSP, with average reductions ranging from 22.85\% to 44.75\%. Moreover, one-sided paired Student’s $t$-tests over all 60 instances confirm that the TS-RSP makespan is significantly smaller than the OS-RSP makespan. Therefore, adopting two-sided shunting can significantly improve yard efficiency and provide greater flexibility in responding to congestion and time-sensitive departures.  Of course, this requires additional potential costs associated with additional track siding space and an additional locomotive.  

\begin{table}[htbp]
\centering
\scriptsize
\setlength{\tabcolsep}{6pt}
\caption{Makespan comparison for OS--RSP and TS--RSP.}
\label{tab:ms_comparison}
\begin{adjustbox}{max width=\textwidth}
\begin{tabular}{lccccc}
\toprule
& \multicolumn{1}{c}{OS--RSP} & \multicolumn{2}{c}{TS--RSP (APS)} & \multicolumn{2}{c}{TS--RSP (ROBS)} \\
\cmidrule(lr){2-2} \cmidrule(lr){3-4} \cmidrule(lr){5-6}
Scale & Makespan & Makespan & MS Red. (\%) & Makespan & MS Red. (\%) \\
\midrule
Small (20)  & 5.95  & 3.75  & 37.32 & 3.50  & 40.89 \\
Medium (20) & 15.20 & 10.60 & 22.85 & 8.75  & 36.30 \\
Large (20)  & 31.30 & 21.35 & 31.84 & 17.15 & 44.75 \\
\bottomrule
\multicolumn{5}{l}{\footnotesize \textit{Note:} The Student's t-tests over all 60 instances yields \(p=6.31\times 10^{-11}\) for APS and \(p=3.66\times 10^{-11}\) for ROBS.}
\end{tabular}
\end{adjustbox}
\end{table}

\section{Conclusion and Future Research}
\label{sec:conclusion}
This paper studies railcar shunting problems in a flatyard.  We consider two potential track layouts, i.e., a One-Sided Railcar Shunting Problem (OS-RSP) and a Two-Sided Railcar Shunting Problem (TS-RSP), where the goal is to construct a sequence of shunting moves that delivers railcar groups to their designated destination tracks while minimizing total shunting cost. The OS-RSP has only one switch end and shunting moves therefore operate according to a LIFO property (similar to a stack), whereas the TS-RSP contains two switch ends and permits using a combination of LIFO and FIFO rules, analogous to a queue. The TS-RSP provides advantages over the OS-RSP by allowing two locomotives to operate simultaneously from two switch ends. To handle the more complex TS-RSP version, we develop decomposition mappings that transform a TS-RSP instance into two coupled OS-RSP subproblems (subproblems $A$ and $B$), which enables parallel decision making while explicitly accounting for coordination between the two switch ends. To the best of our knowledge, this is the first paper considering such settings. The $\mathcal{NP}$-hardness of the OS-RSP optimization problem was established in \textcite{switchingproject}. We generalize this result to the TS-RSP by showing that the OS-RSP serves as a special case of the TS-RSP.

To solve these problems at a practical scale, we propose a novel Hybrid Heuristic-Reinforcement Learning (HHRL) framework that combines domain-driven heuristics, including preprocessing and a fixed $f$-group batching decomposition approach, with a $Q$-learning Model. The preprocessing phase standardizes arbitrary initial yard configurations into a canonical layout and reduces the state-action space by
removing unnecessary groups we defined as tail-home and tail-ready groups, merging groups with a common destination, and deleting unnecessary classification tracks. The fixed $f$-group Batching approach decomposes the resulting canonical layout into a sequence of smaller subproblems, which are then solved individually using $Q$-learning.

Computational experiments on 120 randomly generated instances (60 OS-RSP and 60 TS-RSP) show that the HHRL method produces high-quality solutions. In the OS-RSP tests, for small-scale cases and a subset of medium-scale cases for which optimal solutions are available, the HHRL method achieves low optimality gaps of 3.05\% and 0\%, respectively. For the remaining 14 medium-scale instances (70\% of the medium set), neither the MIP model nor the ARG-DP heuristic returns a solution within the 12-hour time limit, whereas the HHRL method produces feasible solutions with an average runtime of 178.68 seconds. For large-scale cases, the HHRL method generates solutions with an average runtime of 332.68 seconds, indicating an acceptable increase in running time as the instance size grows. In the TS-RSP tests, we compare two decomposition mappings, APS and ROBS. The APS approach biases the workload toward one switch end while the ROBS approach balances the workload between the two switch ends. By comparing their resulting makespan values with those of the OS-RSP, we conclude that the TS-RSP shunts more efficiently, resulting from the ability to coordinate the use of two locomotives at opposite switch ends. 

Future work may extend the shunting problem to settings where departure tracks are not predetermined, allowing all tracks to serve as both classification and departure tracks. Another direction may incorporate stochastic disturbances, where extra railcars may dynamically enter or leave the yard during shunting operations. In addition, incorporating Deep $Q$-Network solution methods may enhance policy learning by better capturing long-term reward structures for problems with large state and action spaces.

\printbibliography




\end{document}